\pdfoutput=1

\documentclass[11pt]{article}

\usepackage[]{emnlp2021}

\usepackage{times}
\usepackage{latexsym}

\usepackage[T1]{fontenc}

\usepackage[utf8]{inputenc}

\usepackage{microtype}

\title{LM-Critic: Language Models for Unsupervised Grammatical Error Correction}

\author{Michihiro Yasunaga \quad Jure Leskovec \quad Percy Liang \\
Stanford University\\
\scalebox{0.87}[0.9]{{\tt \{myasu,jure,pliang\}@cs.stanford.edu}}}

\usepackage{amsmath}
\usepackage{amsfonts}
\usepackage{lipsum} 
\usepackage{algorithm}
\usepackage{graphicx}
\usepackage{transparent}
\usepackage{soul}
\usepackage{calc}
\definecolor{darkred}{HTML}{bb0000}
\usepackage[moderate]{savetrees}

\usepackage{multirow}
\usepackage{makecell}
\usepackage{arydshln}
\usepackage{colortbl}
\usepackage{booktabs} 

\renewcommand\ttdefault{cmtt}
\usepackage{xspace}

\usepackage{amsmath,amsfonts,bm}

\renewcommand{\slash}{\!/\! }
\newcommand{\heading}[1]{\paragraph{#1}}
\newcommand{\eg}{\textit{e.g.}}
\newcommand{\ie}{\textit{i.e.}}
\newcommand{\trainbg}{\textsc{Train}^{\text{bad}\rightarrow \text{good}}}
\newcommand{\traingb}{\textsc{Train}^{\text{good}\rightarrow \text{bad}}}
\newcommand{\goodsent}{x_\text{good}}
\newcommand{\badsent}{x_\text{bad}}
\newcommand{\Fscore}{F$_\text{0.5}$\xspace}

\def\eqref#1{equation~\ref{#1}}

\def\1{\bm{1}}

\DeclareMathAlphabet{\mathsfit}{\encodingdefault}{\sfdefault}{m}{sl}
\SetMathAlphabet{\mathsfit}{bold}{\encodingdefault}{\sfdefault}{bx}{n}

\def\gD{{\mathcal{D}}}

\def\gP{{\mathcal{P}}}

\DeclareMathOperator*{\argmax}{arg\,max}

\usepackage{mathtools}

\begin{document}
\setlength{\abovedisplayskip}{6pt}
\setlength{\belowdisplayskip}{6pt}

\newcommand\pl[1]{\textcolor{red}{[PL: #1]}}

\maketitle

\begin{abstract}

Training a model for grammatical error correction (GEC) requires a set of labeled ungrammatical \slash grammatical sentence pairs, but manually annotating such pairs can be expensive.
Recently, the Break-It-Fix-It (BIFI) framework has demonstrated strong results on learning to repair a broken program without any labeled examples,
but this relies on a perfect critic (\eg, a compiler) that returns whether an example is valid or not, which does not exist for the GEC task.
In this work, we show how to leverage a pretrained language model (LM) in defining an LM-Critic, which judges a sentence to be grammatical if the LM assigns it a higher probability than its local perturbations.
We apply this LM-Critic and BIFI along with a large set of unlabeled sentences to bootstrap realistic ungrammatical \slash grammatical pairs for training a corrector.
We evaluate our approach on GEC datasets across multiple domains (CoNLL-2014, BEA-2019, GMEG-wiki and GMEG-yahoo) and show that it outperforms existing methods in both the unsupervised setting (+7.7 \Fscore) and the supervised setting (+0.5 \Fscore).

\end{abstract}

\section{Introduction}

\begin{figure}[!t]
    \vspace{-4mm}
    \scalebox{0.9}{\textbf{(a)}~ Grammatical error correction (GEC) via LM-Critic}
    \begin{center} \vspace{1mm}
        \includegraphics[width=0.48\textwidth]{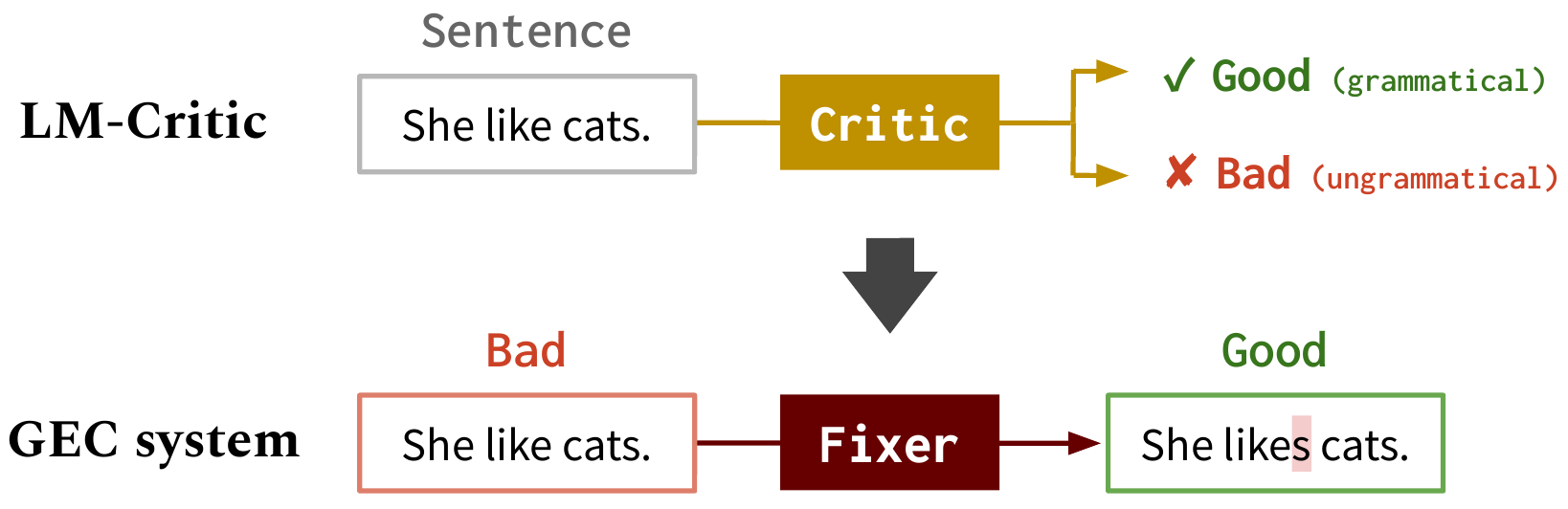}
        \vspace{-1mm}
    \end{center}

    \scalebox{0.9}{\textbf{(b)}~ Idea behind LM-Critic: Local optimum criterion}
    \begin{center}
        \includegraphics[width=0.47\textwidth]{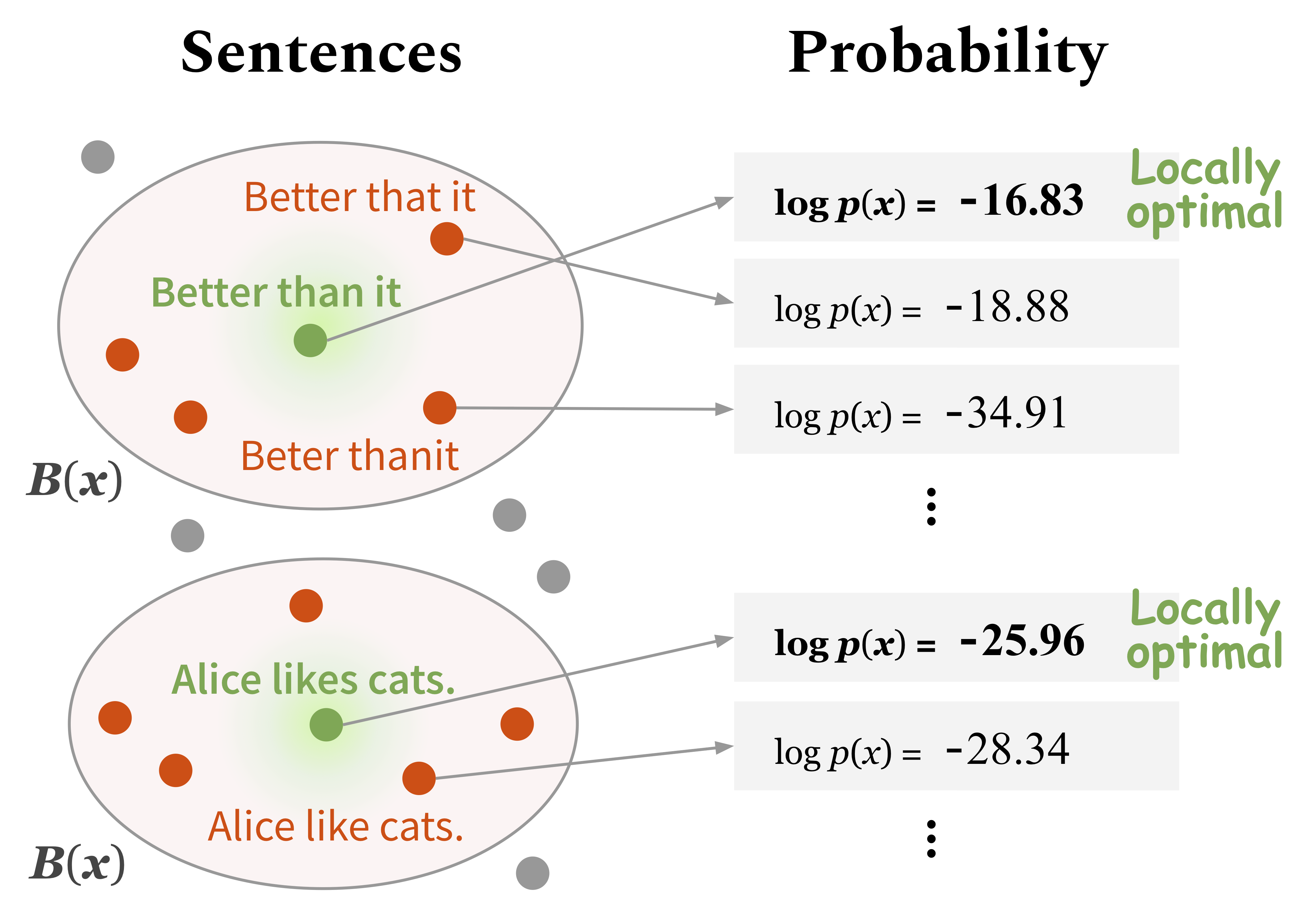}
    \end{center}
    \vspace{-1mm}
    \caption{\small
    Illustration of LM-Critic. \textbf{(a)} In this work, we train a fixer for grammatical error correction (GEC) by leveraging LM-Critic that assesses the grammaticality.
    \textbf{(b)} LM-Critic deems a sentence to be grammatical if a pretrained language model (\eg, GPT2) assigns it a higher probability than candidates in its local neighborhood (\eg, edit distance 1).
    } 
  \label{fig:idea}
\end{figure}

Grammatical error correction (GEC) is the task of fixing grammatical errors in text, such as typos, tense and article mistakes.
Recent works cast GEC as a translation problem, using encoder-decoder models to map bad (ungrammatical) sentences into good (grammatical) sentences \cite{yuan2016grammatical,xie2016neural,ji2017nested,chollampatt2018multilayer,junczys2018approaching}. 
These methods rely on a combination of human-labeled data (\ie, $\langle$bad, good$\rangle$ pairs) \cite{nicholls2003cambridge,yannakoudakis2011new,bryant2019bea} and
synthetic data, which are generated by corrupting good sentences into $\langle$synthetic bad, good$\rangle$ pairs \cite{awasthi2019parallel,kiyono2019empirical}.
Human-labeled pairs are representative of real human errors but are expensive to obtain, while synthetic pairs are cheap but are unrealistic, deviating from the distribution of grammatical errors humans make \cite{grundkiewicz2019neural}. How to obtain inexpensive yet realistic paired data to improve GEC remains a key challenge, especially in domains or languages with no labeled GEC data \cite{napoles2019enabling,naplava2019grammatical}.

Break-It-Fix-It (BIFI; \citet{yasunaga2021break}) is a recent method to obtain realistic paired data from unlabeled data, which has shown promise in the task of source code repair. The idea of BIFI is that using an initial fixer (\eg, trained on synthetic data) and a critic that tells if an input is bad or good (\eg, compiler, which checks if code has an error), BIFI iteratively trains the fixer and a breaker to generate better paired data. Specifically, BIFI (1) applies the fixer to bad examples and keeps outputs \textit{accepted by the critic}, (2) trains a breaker on the resulting paired data and uses it to generate more pairs, and (3) trains the fixer on the pairs generated in Step (1) and (2).
This way, BIFI adapts the fixer to more realistic distributions of $\langle$bad, good$\rangle$ pairs, only using unlabeled data.
However, BIFI is not directly applicable to GEC because it requires an oracle critic (\eg, compiler), which does not exist for GEC.

In this work, we propose \textit{LM-Critic}, a simple approximate critic for assessing grammaticality (\S \ref{sec:lm_critic}), and apply it with BIFI to learn GEC from unlabeled data (\S \ref{sec:gec_learning}).
Specifically, motivated by recent progress in large language models (LMs) (\eg, GPT2, GPT3; \citet{radford2019language, brown2020language}) and an intuition that a good LM assigns a higher probability to grammatical sentences than ungrammatical counterparts, we use an LM’s probability to define a critic for grammaticality. 
A naive approach is to deem a sentence as grammatical if its probability exceeds an absolute threshold, but this does not work in practice, \eg, LMs may assign a high probability just because the sentence has more common words.
We hence compare probabilities in \textit{local} neighborhood of sentences. Concretely, LM-Critic is defined by two components, an LM (\eg, GPT2) and a neighborhood function (\eg, edit distance 1), and deems a sentence to be grammatical if the LM assigns it the highest probability in its local neighborhood (Figure \ref{fig:idea}; local optimum criterion). 
Using this LM-Critic, we apply BIFI to the GEC task.
Notably, our approach, both the LM-Critic and GEC learning, does not require labeled data.

We evaluate our proposed approach on GEC benchmarks across multiple domains, CoNLL-2014 \cite{ng2014conll}, BEA-2019 \cite{bryant2019bea}, GMEG-yahoo, and GMEG-wiki \cite{napoles2019enabling}.
We achieve strong performance in the unsupervised setting (\ie, no labeled data), outperforming the baseline fixer trained on synthetic data by 7.7 \Fscore on average. 
We also evaluate in the supervised setting, where we take the state-of-the-art model GECToR \cite{omelianchuk2020gector} as the baseline fixer, and further fine-tune it by applying our approach using unlabeled data. We achieve 65.8 \slash 72.9 \Fscore on CoNLL-2014 \slash BEA-2019, outperforming GECToR by 0.5 \Fscore. 
Our results also suggest that while existing BIFI assumed access to an oracle critic (\ie, compiler), an approximate critic (\ie, LM-Critic) can also help to improve model learning.
\section{Problem setup}

The task of grammatical error correction (GEC) is to map an ungrammatical sentence $\badsent$ into a grammatical version of it, $\goodsent$ (one that has the same intended meaning). 
A GEC model (\textit{fixer}) $f$ aims to learn this mapping, typically using a \textit{paired} dataset $\gD_{\text{pair}} = \{({\badsent}^{(i)}, {\goodsent}^{(i)})\}$.
In particular, we call it \textit{labeled} if the pairs are human-annotated. 
In contrast, we call \textit{unlabeled} data a set of raw sentences $\gD_{\text{unlabel}} = \{x^{(i)}\}$.
For simplicity, we use ``good'' \slash ``bad'' to mean grammatical \slash ungrammatical interchangeably.
Unlike a fixer, which maps $\badsent$ to $\goodsent$, a \textit{critic} $c$ merely assesses whether an input is good or bad: for a sentence $x$,
\begin{align}\vspace{-1mm}
    c(x) = \begin{cases}
    1 ~~~\text{if $x$ is good}\\[-1mm]
    0 ~~~\text{if $x$ is bad}.
    \end{cases}
    \vspace{-1mm}
\end{align}

Given unlabeled data $x$'s (some of which are good, some of which are bad), and a language model (LM), which returns a probability distribution $p(x)$ over sentences $x$, we aim to define the critic (\S \ref{sec:lm_critic}; LM-Critic) and use that to obtain the fixer (\S \ref{sec:gec_learning}; BIFI).

\section{LM-Critic}
\label{sec:lm_critic}

The core of our approach to GEC is a critic, which returns whether a sentence is good (grammatical) or bad (ungrammatical).
Motivated by recent progress in large-scale pre-trained LMs (\eg, GPT2, GPT3; \citet{radford2019language,brown2020language}), we aim to use an LM’s probability score to define a critic for grammaticality.
Specifically, we propose a criterion that deems a sentence to be good if it has the highest probability within its local neighborhood (local optimum criterion; \S \ref{sec:local_optimum}).
We implement this criterion using a pretrained LM and a sentence perturbation function (LM-Critic; \S \ref{sec:lm_critic_implementation}).
We then do an intrinsic study on how well LM-Critic works in practice (\S \ref{sec:lm_critic_analysis}).

\subsection{\scalebox{0.97}[1]{Local optimum criterion of grammaticality}}
\label{sec:local_optimum}

Our starting point is the idea that a good LM assigns a higher probability to grammatical sentences than ungrammatical ones.
With this idea, a naive way to judge grammaticality might be to find a threshold ($\delta$) for the absolute probability, and let the critic be:
\begin{align}\vspace{-1mm}
    \text{AbsThr-Critic}(x) = \begin{cases}
    1 ~~~\text{if ~${p}(x) > \delta$}\\[-1mm]
    0 ~~~\text{otherwise}.\label{eq:abs_thr}
    \end{cases}
    \vspace{-1mm}
\end{align}
However, this does not work in practice. In Figure \ref{fig:idea}, for instance, ``\textit{Alice likes cats}'' (4th sentence) is grammatical but has a lower probability (according to GPT2) than  ``\textit{Better that it}'' (2nd sentence), which is ungrammatical. This is because the two sentences have different meanings and are not directly comparable. We also empirically find that this critic based on absolute threshold does not work well (\S \ref{sec:lm_critic_analysis_performance}).

This observation motivates us to compare sentences with the same intended meaning, and leads to the following two refined intuitions.

\heading{Intuition 1 (Correlation of grammaticality and probability).} For a grammatical sentence, $\goodsent$, and an ungrammatical \textit{version} of it (with the same intended meaning), $\badsent$, we have
\begin{align}
    p(\badsent) < p(\goodsent).
\end{align}

\heading{Intuition 2 (Local neighborhood of sentences).}
Assume for simplicity that every sentence has exactly one grammatical version of it (\ie, if the sentence is grammatical, itself; if not, its corrected version).\footnote{We acknowledge that this assumption may not hold in some cases, \eg, an ungrammatical sentence may have no correction (``asdfghgfdsa''---just a random typo?) or multiple corrections (``The cat sleep.''---change ``sleep'' to the present tense or past?). We accept this assumption considering that it is often sufficient in common GEC datasets, and leave the relaxation of the assumption for future work.\vspace{-0mm}}
For each sentence $x$, there is a set of sentences, $B(x)$ (\textit{local neighborhood}), that consists of the grammatical version and all other ungrammatical versions of $x$.\vspace{3mm}

\noindent
Assuming the above two intuitions, we obtain the following criterion for judging grammaticality, where the idea is to compare sentences within the meaning-preserving local neighborhood.

\heading{Local optimum criterion of grammaticality.}
For each sentence $x$, we let $B(x)$ be its local neighborhood as defined in Intuition 2. We then have
\begin{align}
   \text{$x$ is grammatical ~iff~}~ x = \argmax_{x'\in B(x)}~ p(x'). \label{eq:local_optimum}
\end{align}

\noindent
The justification is as follows. If $x$ is grammatical, then by Intuition 1, $x$ has a higher probability than any other sentences in $B(x)$, as they are ungrammatical; hence, we have the RHS of iff. On the other hand, if $x$ is ungrammatical, then by Intuition 1, the grammatical version of $x$ has a higher probability than $x$, which contradicts with the RHS of iff.

The idea is to deem a sentence to be grammatical if it has the highest probability within its meaning-preserving local neighborhood (Figure \ref{fig:idea}).
We will next describe how to implement this criterion in practice.

\subsection{Implementation of LM-Critic}
\label{sec:lm_critic_implementation}
We implement LM-Critic by approximating the local optimum criterion.
First, for the sentence probability $p(x)$, we use a pretrained LM's probability score.
As obtaining the ground-truth local neighborhood $B(x)$ is difficult, we aim to get an approximate, $\hat B(x)$: we implement a sentence perturbation function $b$, and let $\hat B(x)$ be samples from $b(x)$.
To check the grammaticality of a sentence, we apply the local optimum criterion (Eq \ref{eq:local_optimum}) using $\hat{B}(x)$:
\begin{align}
\vspace{-1mm}
\scalebox{1}{\text{$
    \text{LM-Critic}(x) = \begin{cases}
    1 ~~~\text{if $\displaystyle x = \argmax_{x'\in \hat{B}(x)}~ p(x')$}\!\!\!\!\!\!\\[-1mm]
    0 ~~~\text{otherwise}.
    \end{cases}
    $}}
    \vspace{-1mm}
\end{align}

There are three decisions for implementing LM-Critic: choice of a pretrained LM, perturbation function $b$, and sampling method of perturbations.

\heading{Pretrained LM.} We experiment with various sizes of GPT2 models \cite{radford2019language}---GPT2 (117M parameters), GPT2-medium (345M), GPT2-large (774M), GPT2-xl (1.6B). These LMs were trained on a large set of web text (40GB).

\heading{Perturbation function.}  We study three variants:
\begin{itemize}
    \setlength{\itemsep}{1mm}
    \item[$\bullet$] \textbf{ED1.~} Given a sentence, we generate edit-distance one (ED1) perturbations in the character space. Following prior works in typo generation \cite{pruthi2019combating,jones2020robust}, we randomly insert a lowercase letter, delete a character, replace a character, or swap two adjacent characters.

    \item[$\bullet$] \textbf{ED1 + Word-level heuristics (all).~} ED1 can cover most of the character-level typos but may not cover word-level grammatical errors, such as missing an article. Besides ED1, here we include heuristics for word-level perturbations used in \citet{awasthi2019parallel}, which randomly inserts, deletes, or replaces a word based on its dictionary. Please refer to \citeauthor{awasthi2019parallel} for more details.

    \item[$\bullet$] \textbf{ED1 + Word-level heuristics.~}  We noticed that the above word-level heuristics include perturbations that may alter the meaning of the original sentence (\eg, deleting \slash inserting ``not'').
    Therefore, we remove such heuristics here.
\end{itemize}

\heading {Sampling perturbations.}
As the output space of the perturbation function $b$ is large, we obtain samples from $b(x)$ to be $\hat{B}(x)$.
We experiment with random sampling with sizes of 100, 200 and 400, motivated by the finding that with the GPT2 models, a batch size of 100 sentences can fit into a single GPU of 11GB memory.
Other (potentially more efficient) sampling methods include gradient-based sampling which picks perturbation sentences in a direction that increases the sentence probability (analogous to adversarial perturbations; \citet{szegedy2013intriguing,wallace2019universal}), but we focus on random sampling in this work.

\heading{}
The advantage of LM-Critic is that as LMs can be trained on a wide range of unlabeled corpora, it is unsupervised and usable in various domains of text.

\subsection{Empirical analysis}
\label{sec:lm_critic_analysis}

We study how well our LM-Critic works in practice.
We prepare an evaluation data for judging grammaticality in \S \ref{sec:lm_critic_analysis_data}. We first perform a simple check to make sure that LMs' probability score correlates with grammaticality (\S \ref{sec:lm_critic_analysis_likelihood}).
We then study the performance of LM-Critic judging grammaticality (\S \ref{sec:lm_critic_analysis_performance}).
The analysis we conduct in this section is just an intrinsic evaluation of LM-Critic. Our main goal is to use LM-Critic with BIFI for learning GEC, which we describe and evaluate in \S \ref{sec:gec_learning}.

\subsubsection{Evaluation data}
\label{sec:lm_critic_analysis_data}
To gain insights into how well LM-Critic judges grammaticality, we prepare a simple evaluation data consisting of $(\badsent, \goodsent)$ sentence pairs.
As experimenting with multiple datasets is desired in GEC \cite{mita2019cross}, we construct a combined evaluation set from the dev sets of multiple GEC benchmarks, GMEG-wiki \cite{napoles2019enabling}, GMEG-yahoo, and BEA-2019 \cite{bryant2019bea}, which span the domains of Wikipedia, Yahoo!\! Answers, and essay \slash learner English.
Specifically, we sampled $\sim$600 labeled pairs of $(\badsent, \goodsent)$ in total from the three benchmarks. We filter out examples where $\badsent = \goodsent$ in this process.
We acknowledge that while we use annotated $(\badsent, \goodsent)$ pairs for the evaluation here, this does not fully match the way LM-Critic will be used in BIFI (\S \ref{sec:gec_learning}), where the critic is run on unlabeled sentences; our study here is just to gain intrinsic insights into LM-Critic.

\begin{figure}[!t]
    \centering
    \vspace{-2mm}
    \includegraphics[width=0.25\textwidth]{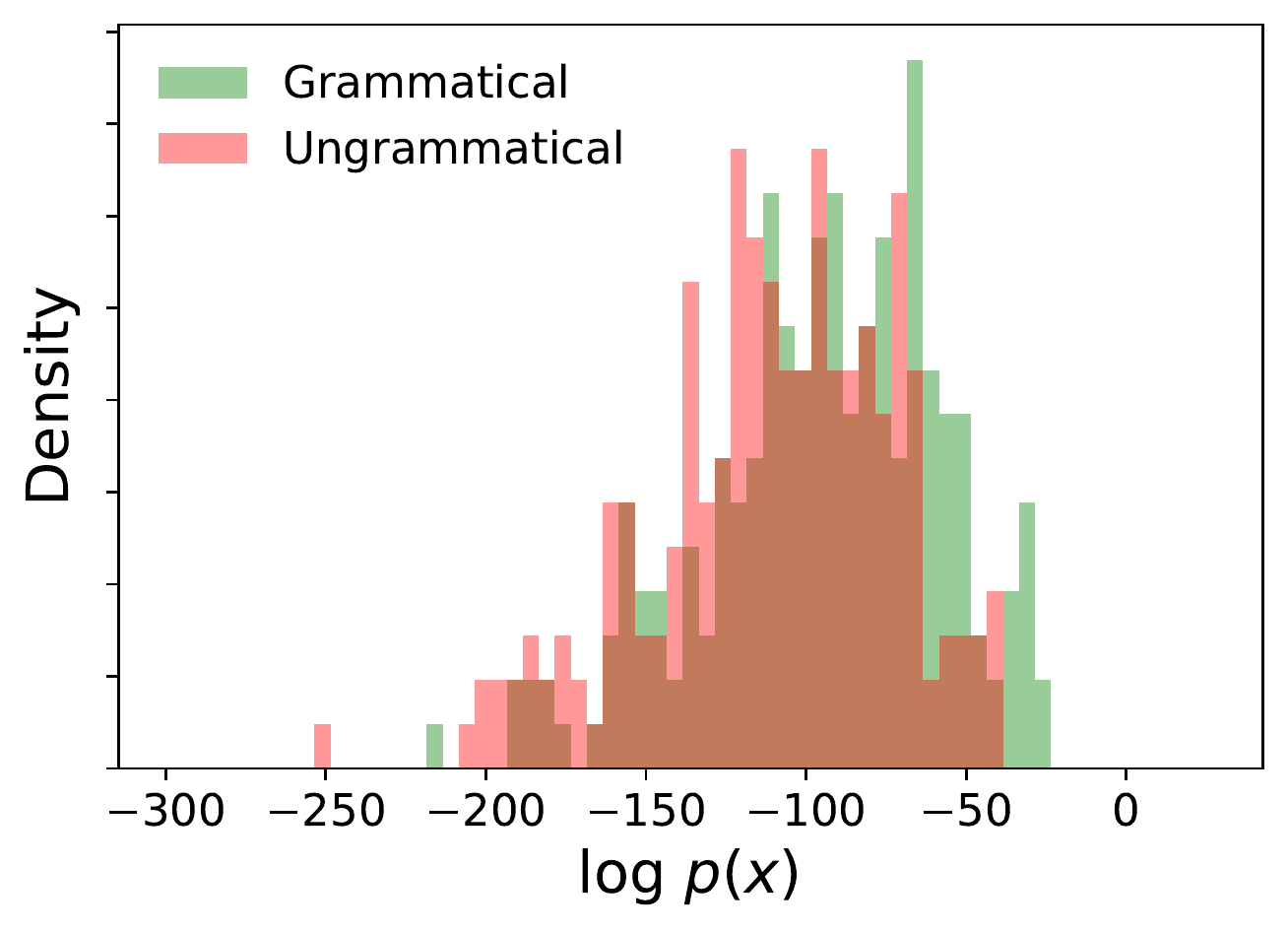}
    \vspace{-3mm}
    \caption{\small
    Probability of grammatical (green) and ungrammatical (red) sentences, computed by a pretrained LM (GPT2).
    }
  \label{fig:histogram}
\end{figure}

\begin{table}[tb]
\centering
\scalebox{0.65}{
\begin{tabular}{lc}
\toprule
\textbf{Pretrained LM}~~ & \textbf{How often $\boldsymbol{p(\badsent) < p(\goodsent)}$?} \\
\midrule
GPT2 &  94.7\% \\
GPT2-medium & 95.0\% \\
GPT2-large & 95.9\% \\
GPT2-xl & 96.0\% \\
\bottomrule
\end{tabular}
}
\vspace{-2mm}
\caption{\small
How well sentence probability returned by pretrained LMs correlates with grammaticality empirically.
}
\label{tab:lm_likelihood}
\end{table}

\subsubsection{Analysis of LM probability}
\label{sec:lm_critic_analysis_likelihood}
Using the evaluation data, we first make sure that pretrained LMs' probability correlates with grammaticality.
Figure \ref{fig:histogram} shows a histogram for the probability $\log p(x)$ of grammatical (green) and ungrammatical (red) sentences computed by GPT2.
In Table \ref{tab:lm_likelihood}, we study how often pretrained LMs actually assign a higher probability to $\goodsent$ than $\badsent$ on the evaluation pairs $(\badsent, \goodsent)$.
We find that the LMs satisfy $p(\badsent) < p(\goodsent)$ about 94\% of the time, with a slight increase when using a larger model (from GPT2 to GPT2-xl).
We find that the remaining pairs with $p(\badsent) > p(\goodsent)$ consist mostly of cases where $\goodsent$ adds commas or quotations to $\badsent$ (see Table \ref{tbl:lm_critic_error} top for examples).

\begin{table}[!t]
\centering
\scalebox{0.65}{
\begin{tabular}{lcccccc}
    \toprule  
    \multirow{2}{*}{\textbf{Perturbation}\vspace{-2mm}}
    & \multicolumn{3}{c}{\textbf{Recognize ``Good''}} & \multicolumn{3}{c}{\textbf{Recognize ``Bad''}}\\
    \cmidrule(lr){2-4} \cmidrule(lr){5-7}
      & \textbf{P} & \textbf{R} & \textbf{F$_\text{0.5}$} & \textbf{P} & \textbf{R} & \textbf{F$_\text{0.5}$}\\
    \midrule  
    ED1 & 58.7 & 90.1 & 63.1 & 78.8 & 36.8 & 64.2 \\
    ED1 + word(all) & 69.7 & 10.2 & 32.2 & 51.5 & 95.5 & 56.7\\
    ED1 + word &  68.4 & 75.5 & \textbf{69.7} & 72.7 & 65.1 & \textbf{71.1}\\
    \bottomrule 
\end{tabular}
}

\begin{center}
    \vspace{1mm}
    \scalebox{0.65}{
    \begin{tabular}{lcccccc}
        \toprule  
        \multirow{2}{*}{\textbf{Sample size}\vspace{-2mm}}
        & \multicolumn{3}{c}{\textbf{Recognize ``Good''}} & \multicolumn{3}{c}{\textbf{Recognize ``Bad''}}\\
        \cmidrule(lr){2-4} \cmidrule(lr){5-7}
          & \textbf{P} & \textbf{R} & \textbf{F$_\text{0.5}$} & \textbf{P} & \textbf{R} & \textbf{F$_\text{0.5}$}\\
        \midrule  
        100 &  68.4 & 75.5 & {69.7} & 72.7 & 65.1 & {71.1}\\
        200 & 71.3 & 71.5 & 71.4 & 71.4 & 71.3 & \textbf{71.4}\\
        400 &  72.6 & 68.7 & \textbf{71.8} & 70.3 & 74.0 & 71.0\\
        \bottomrule 
    \end{tabular}
    }
\end{center}

\begin{center}
    \vspace{1mm}
    \scalebox{0.65}{
    \begin{tabular}{lcc}
        \toprule  
        \multirow{2}{*}{\textbf{Pretrained LM}\vspace{-2mm}}
        & \multicolumn{1}{c}{\textbf{Recognize ``Good''}} & \multicolumn{1}{c}{\textbf{Recognize ``Bad''}}\\
        \cmidrule(lr){2-2} \cmidrule(lr){3-3}
          & \textbf{F$_\text{0.5}$} & \textbf{F$_\text{0.5}$}\\
        \midrule  
        GPT2 & 69.7 & 71.1 \\
        GPT2-medium & 69.9 & 71.0\\
        GPT2-large  & {70.3} & 71.3\\
        GPT2-xl  & 69.9 & {71.0}\\
        \bottomrule 
    \end{tabular}
    }
\end{center}
\vspace{-3mm}
\caption{\small
\textbf{Performance of LM-Critic}, when using different choices of a perturbation function, sample size, and pretrained LM described in \S \ref{sec:lm_critic_implementation}.
\textbf{(Top)} We set the LM to be GPT2 and the perturbation sample size to be 100, and vary the perturbation function $b$. ``ED1 + word'' achieves the best F$_\text{0.5}$. Henceforth, we use this perturbation function.
\textbf{(Middle)} We set the LM to be GPT2 and vary the perturbation sample size. Increasing the sampling size improves the performance slightly.
\textbf{(Bottom)} We vary the LM. Increasing the LM size makes slight or no improvement in \Fscore on the dataset we used.
}
\label{tbl:lm_critic}
\end{table}

\begin{table}[tb]
\definecolor{lightyellow}{HTML}{FFF7BF}
\definecolor{lightred}{HTML}{f4d7d7} 
\definecolor{lightblue}{HTML}{DAE8FC}
\newcommand{\hlyellow}[1]{{\sethlcolor{yellow}\hl{#1}}}
\newcommand{\hlblue}[1]{{\sethlcolor{lightblue}\hl{#1}}}
\hspace{-2mm}
\scalebox{0.7}{
\begin{tabular}{l}
\toprule
\textbf{Examples of~~ $\boldsymbol{p(\badsent) > p(\goodsent)}$} \\
\midrule
{\small\bf (Comma)}\\[-1mm]
\hspace{1mm}$\badsent$:\hspace{1.5mm} {\small The video was filmed on January 22 and is set to premiere on February 22.}\\[-1mm]
\hspace{1mm}$\goodsent$: {\small The video was filmed on January 22\hlyellow{,~} and is set to premiere on February 22.}\\
{\small\bf (Quotation)}\\[-1mm]
\hspace{1mm}$\badsent$:\hspace{1.5mm} {\small Uprising is a 1980 roots reggae album by Bob Marley \& The Wailers.}\\[-1mm]
\hspace{1mm}$\goodsent$: {\small \hlyellow{``}Uprising\hlyellow{''} is a 1980 roots reggae album by Bob Marley \& The Wailers.}\\
{\small\bf (British spelling)}\\[-1mm]
\hspace{1mm}$\badsent$:\hspace{1.5mm} {\small The blast could be heard across the whole city centre.}\\[-1mm]
\hspace{1mm}$\goodsent$: {\small The blast could be heard across the whole city \hlyellow{center}.}\\[2mm]
\toprule
\textbf{Examples of~~ $\boldsymbol{p(x')  > p(\goodsent), ~x' \in \hat{B}(\goodsent)}$} \\
\midrule
{\small\bf (Singular \slash plural)}\\[-1mm]
\hspace{1mm}$x'$:\hspace{4.8mm} {\small They are affiliated to either the state board\hlyellow{s} or to national education boards.}\\[-1mm]
\hspace{1mm}$\goodsent$: {\small They are affiliated to either the state board or to national education boards.}\\
{\small\bf (Tense)}\\[-1mm]
\hspace{1mm}$x'$:\hspace{4.8mm} {\small As well as touring Europe, they tour with such acts as Green Day.}\\[-1mm]
\hspace{1mm}$\goodsent$: {\small As well as touring Europe, they tour\hlyellow{ed} with such acts as Green Day.}\\[0.5mm]

\bottomrule
\end{tabular}
}
\vspace{-2mm}
\caption{\small
Failure cases of LM-Critic. (Top) GPT2 assigns a higher probability to bad sentences.
(Bottom) our neighborhood function (``ED1 + word'') includes sentences with a higher LM probability than the original good sentence.
}
\label{tbl:lm_critic_error}
\end{table}

\subsubsection{Performance of LM-Critic}
\label{sec:lm_critic_analysis_performance}

In \S \ref{sec:lm_critic_analysis_likelihood} we simply made sure that pretrained LMs' probability correlates with grammaticality.
Here we study LM-Critic's performance of \textit{judging} bad \slash good sentences, on the evaluation set $\{({\badsent}^{(i)}, {\goodsent}^{(i)})\}$.
We treat the label of $\badsent$'s and $\goodsent$'s to be ``bad'' and ``good'', respectively, and measure the precision (P), recall (R), \Fscore of LM-Critic recognizing ``bad'' and ``good''. Denoting the critic as $c$, precision and recall for ``bad'' are defined as
\begin{align}
    \scalebox{0.9}{\text{$\text{P}^\text{(bad)}$}} = \scalebox{0.85}{\text{$\displaystyle\frac{|\{x\!:~ c(x)=0 \}| ~\cap~ |\{\badsent\}|}{|\{x\!:~ c(x)=0 \}|}$}}, \\
    \scalebox{0.9}{\text{$\text{R}^\text{(bad)}$}} = \scalebox{0.85}{\text{$\displaystyle\frac{|\{x\!:~ c(x)=0 \}| ~\cap~ |\{\badsent\}|}{|\{\badsent\}|}$}}.
\end{align}
$\text{P}^\text{(good)}$ and $\text{R}^\text{(good)}$ are defined similarly.
F$_{0.5}$ score is a combined metric of P and R that is commonly used in grammatical error detection \slash correction literature.

\heading{Baseline critic.}
First, as a baseline, we evaluate the critic based on absolute threshold, described in Eq \ref{eq:abs_thr}.
We set the threshold $\delta$ as the average probability of all good and bad sentences in the evaluation data.
This method achieves 54.3 \Fscore\!\!$^\text{(bad)}$ and 56.0 \Fscore\!\!$^\text{(good)}$, using GPT2.

\heading{Proposed LM-Critic.}
Table \ref{tbl:lm_critic} shows the results of our proposed LM-Critic, using different choices of a perturbation function, sample size, and pretrained LM.
Recall that LM-Critic predicts ``bad'' correctly if it finds a perturbed sentence with higher probability, and predicts ``good'' correctly if the input has the highest probability among the sampled perturbations.

\begin{itemize}
    \setlength{\itemsep}{1mm}
    \item[$\bullet$] \textbf{Perturbation function $\boldsymbol{b}$ \text{(top table)}.}
    We set the pretrained LM to be GPT2 and the perturbation sample size to be 100, and vary the perturbation function. We find that when the perturbation space is small (``ED1''), LM-Critic may make false predictions of ``good'', leading to low $\text{P}^\text{(good)}$ and low $\text{R}^\text{(bad)}$. When the  perturbation space is large (``ED1 + word(all)''), LM-Critic may make false predictions of ``bad'', leading to low  $\text{R}^\text{(good)}$ and low $\text{P}^\text{(bad)}$.
    ``ED1 + word'' is the most balanced and achieves the best F$_\text{0.5}$; henceforth, we use this perturbation method for all our experiments.
    Overall, our LM-Critic outperforms the baseline critic by substantial margins.

    \item[$\bullet$] \textbf{Sample size of perturbations \text{(middle table)}.}
    We set the LM to be GPT2 and vary the perturbation sample size. Increasing the sample size tends to improve $\text{P}^\text{(good)}$ and $\text{R}^\text{(bad)}$, and improve the overall F$_\text{0.5}$ performance slightly.

    \item[$\bullet$] \textbf{Pretrained LM \text{(bottom table)}.} We vary the LM. Increasing the LM size makes slight or no improvement in \Fscore on the dataset we used.
\end{itemize}

\heading{}
We also analyze when LM-Critic fails.
When LM-Critic predicts a false ``good'' (labeled ``bad'' but predicted ``good''), it is commonly because of $p(\badsent) > p(\goodsent)$ (as described in \S \ref{sec:lm_critic_analysis_likelihood}; Table \ref{tbl:lm_critic_error} top), or perturbation sampling not hitting a better version of the input $\badsent$.
When LM-Critic predicts a false ``bad'' (labeled ``good'' but predicted ``bad''), it is because some perturbation $x' \in \hat{B}(\goodsent)$ yields $p(x') > p(\goodsent)$. Common examples are the change of tense or singular \slash plural (see Table \ref{tbl:lm_critic_error} bottom for examples).
This indicates that even if we use a conservative edit-distance like ED1, there may be unnecessary perturbations (tense, singular \slash plural) that pretrained LMs prefer, which is a limitation of our current LM-Critic.

The analysis done in this section is an intrinsic evaluation of LM-Critic. Our main goal is to use LM-Critic with BIFI for learning GEC, which we describe in \S \ref{sec:gec_learning}.
While LM-Critic is not perfect in itself as we have seen in this section (it is an \textit{approximate} critic), we will show that it is helpful for obtaining realistic paired data to improve the downstream GEC performance.
Henceforth, we use the ``ED1 + word'' perturbation, a sample size of 100, and GPT2 for our LM-Critic.

\section{Learning GEC with LM-Critic}
\label{sec:gec_learning}

Break-It-Fix-It (BIFI; \citet{yasunaga2021break}) is an existing method that uses a critic to obtain realistic paired data from unlabeled data. BIFI was originally studied in the source code repair task where an oracle critic (\eg, compiler) exists, but there is no oracle critic in GEC. 
Here, we propose to apply BIFI to the GEC task by using LM-Critic as the critic (\S \ref{sec:gec_method}), and evaluate this approach on GEC benchmarks (\S \ref{sec:gec_exp}).
The difference from the original BIFI is that our task is GEC rather than code repair, and we use an approximate critic (\ie, LM-Critic) instead of an oracle critic (\ie, compiler).

\subsection{Approach}
\label{sec:gec_method}

Our goal is to learn a fixer $f$ that maps an ungrammatical sentence $\badsent$ into the grammatical version $\goodsent$. 
A common method to obtain paired data for GEC from unlabeled text is to heuristically corrupt good sentences (synthetic data) \cite{awasthi2019parallel,kiyono2019empirical}. However, such synthetic errors do not match the distributions of real grammatical errors humans make, which may result in accuracy drops \cite{daume2006domain}.
To mitigate this mismatch, BIFI aims to obtain more realistic paired data and train the fixer on it. 

Specifically, BIFI takes as inputs:
\begin{itemize}
    \item[$\bullet$] \textbf{Critic} $c$, for which we use LM-Critic
    
    \item[$\bullet$] \textbf{Unlabeled data} $\gD_\text{unlabel}$. Using the critic $c$, examples in $\gD_\text{unlabel}$ can be split into {bad} ones $\gD_\text{bad} = \{x \mid x \in \gD_\text{unlabel},~ c(x) = 0 \}$ and {good} ones $\gD_\text{good} = \{y \mid y \in \gD_\text{unlabel},~  c(y) = 1 \}$
    
    \item[$\bullet$] \textbf{Initial fixer} $f_0$, which could be trained on synthetic data (unsupervised setting; \S \ref{sec:gec_exp_low_resource}) or labeled data (supervised setting; \S \ref{sec:gec_exp_full_resource})
\end{itemize}
and improves the fixer by performing a cycle of data generation and training: (1) we
apply the fixer $f$ to the bad examples $\gD_\text{bad}$, which consists of real grammatical errors made by humans, and use the critic to assess if the fixer's output is good---if good, we keep the pair; (2) we train a \textit{breaker} $b$ on the resulting paired data---consequently, the breaker can generate more realistic errors than the initial synthetic data; (3) we apply the breaker to the good examples $\gD_\text{good}$; (4) we finally train the fixer on the newly-generated paired data in (1) and (3).
This cycle can be iterated to improve the fixer and the breaker simultaneously.  
Formally, BIFI does the following in each round $k$ ($=1,2,..., K$):
\begin{align}
\scalebox{0.91}{\text{$\gP_k^{(f)}$}} &= \scalebox{0.91}{\text{$\{ (x,~ f_{k\!-\!1}(x)) \mid x \in \gD_\text{bad},~ \textcolor{darkred}{c(f_{k\!-\!1}(x)) = 1} \}$}} \label{eq:BIFI-Pf}\\
\scalebox{0.91}{\text{$b_k$}} &= \scalebox{0.91}{\text{$\traingb ( \gP_k^{(f)} )$}} \label{eq:BIFI-b}\\
\scalebox{0.91}{\text{$\gP_k^{(b)}$}} &= \scalebox{0.91}{\text{$\{ (b_{k}(y),~ y)  \mid y \in \gD_\text{good},~ \textcolor{darkred}{c(b_{k}(y)) = 0} \}$}} \label{eq:BIFI-Pb}\\
\scalebox{0.91}{\text{$f_k$}} &= \scalebox{0.91}{\text{$\trainbg ({\gP_k^{(f)}~\cup}~ \gP_k^{(b)} )$}},\label{eq:BIFI-f}
\end{align}
where each equation corresponds to the steps (1)--(4) in the description above.
$\traingb(\gP)$ trains an encoder-decoder model that maps ``good''-side examples to ``bad''-side examples in paired data $\gP$, and $\trainbg(\gP)$ does the reverse.
\textcolor{darkred}{\textbf{Red font}} indicates the use of critic.
The key intuition of BIFI is that thanks to the critic,  (i) we can extract $\gD_\text{bad}$ from the unlabeled data $\gD_\text{unlabel}$ and incorporate realistic grammatical errors into our data (as opposed to the synthetic data), and (ii) we can verify if the ``bad''-side and ``good''-side of the generated pairs are actually ``bad'' and ``good'' (Eq \ref{eq:BIFI-Pf}, \ref{eq:BIFI-Pb}; \textcolor{darkred}{\textbf{red font}}), which improves the correctness of generated training data compared to vanilla backtranslation \cite{sennrich2015improving,lample2017unsupervised}.
We refer readers to \citet{yasunaga2021break} for more details.

\begin{table*}[h!]
\vspace{-5mm}
\centering
\scalebox{0.7}{
\begin{tabular}{l ccc ccc ccc ccc}
\toprule
\multirow{2}{*}{\textbf{GEC system}\vspace{-2mm}} & \multicolumn{3}{c|}{\textbf{CoNLL-2014 (test)}} & \multicolumn{3}{c}{\textbf{BEA-2019 (dev)}} & \multicolumn{3}{c}{\textbf{GMEG-wiki (test)}} & \multicolumn{3}{c}{\textbf{GMEG-yahoo (test)}} \\ \cmidrule(lr){2-4} \cmidrule(lr){5-7} \cmidrule(lr){8-10} \cmidrule(lr){11-13} 
                       & \textbf{P} &  \textbf{R}  & \textbf{\Fscore} & \textbf{P} &  \textbf{R}  & \textbf{\Fscore} & \textbf{P} &  \textbf{R}  & \textbf{\Fscore} & \textbf{P} &  \textbf{R}  & \textbf{\Fscore} \\
\midrule
Transformer &
    59.2 & 29.2 & 49.1 &
    44.2 & 17.9 & 34.1 &
    52.1 & 26.5 & 43.7 &
    44.4 & 36.9 & 42.7 \\
+ BIFI with no critic &
    58.2 & 29.9 & 48.9 &
    43.8 & 18.7 & 34.5 &
    53.5 & 27.4 & 44.9 &
    45.1 & 38.5 & 43.6\\
+ \textbf{BIFI (ours)} &
\textbf{64.4} & \textbf{35.6} & \textbf{55.5} &
\textbf{51.6} & \textbf{24.7} & \textbf{42.4} & 
\textbf{57.9} & \textbf{33.6} & \textbf{50.6} & \textbf{53.7} & \textbf{47.1} & \textbf{52.2}\\
\bottomrule
\end{tabular}
}\vspace{-2mm}
\caption{\small
GEC results in the \textbf{unsupervised setting} (\S \ref{sec:gec_exp_low_resource}).
``Transformers'' is trained on synthetic paired data as in \citet{awasthi2019parallel}. If we train it on more realistic paired data generated by BIFI (bottom row), it achieves improved results.
}
\label{tbl:result_unlabeled}
\end{table*}

\begin{table*}
\begin{minipage}{.6\textwidth}
\centering
\scalebox{0.6}{
\begin{tabular}{lc|ccc|ccc}
\toprule
\multirow{2}{*}{\textbf{GEC system}\vspace{-2mm}} & \multirow{2}{*}{\textbf{Ens.}\vspace{-2mm}}& \multicolumn{3}{c|}{\textbf{CoNLL-2014 (test)}} & \multicolumn{3}{c}{\textbf{BEA-2019 (test)}} \\ \cmidrule(lr){3-5} \cmidrule(lr){6-8} 
                        & & \textbf{P} &  \textbf{R}  & \textbf{\Fscore} & \textbf{P} &  \textbf{R}  & \textbf{\Fscore}  \\ \midrule
GPT3 (175B) with prompting &&  62.4  &  25.0   &   48.0   &   50.8  &  38.2   & 47.6  \\ 
\midrule
\citet{zhao2019improving}&&   67.7   &  40.6   &   59.8   &   -   &  -   &   -  \\ 
\citet{awasthi2019parallel}&&   66.1   &  43.0    &   59.7   &   -   &  -    &   -  \\ 
\citet{kiyono2019empirical}&&   67.9  &  \textbf{44.1}   &  61.3  & 65.5 &  \textbf{59.4}   &   64.2   \\ \midrule
\citet{zhao2019improving}&\checkmark&   74.1   &  36.3   &  61.3   &   -   &  -   &   -   \\ 
\citet{awasthi2019parallel}&\checkmark&   68.3   &  43.2    &   61.2   &   -   &  -    &   -  \\ 
\citet{grundkiewicz2019neural}&\checkmark&  -  & -   &  64.2  & 72.3 &  60.1   &   69.5    \\ 
\citet{kiyono2019empirical}&\checkmark&   72.4  &  46.1   &  65.0  & 74.7 &  56.7   &   70.2    \\ 
\citet{kantor2019learning}&\checkmark&   - &  -   &  -  & 78.3 &  58.0   &   73.2   \\ 
\midrule
GECToR \cite{omelianchuk2020gector} & & 77.5 &  40.1  & 65.3 & 79.2  & 53.9 & 72.4   \\
\midrule
GECToR ({our base}) & & 77.5 &  40.1  & 65.3 & 79.2  & 53.9 & 72.4  \\
+ \textbf{BIFI (ours)} & & \textbf{78.0} & 40.6 & \textbf{65.8} & \textbf{79.4} & 55.0 & \textbf{72.9} \\
\bottomrule
\end{tabular}
}\vspace{-2mm}
\caption{\small
GEC results in the \textbf{supervised setting} with labeled data available (\S \ref{sec:gec_exp_full_resource}). ``Ens.'' indicates an ensemble system.
}
\label{tbl:result_leaderboard}
\end{minipage}\hfill
\begin{minipage}{.35\textwidth}
    \begin{center}\vspace{1mm}
        \includegraphics[width=0.99\textwidth]{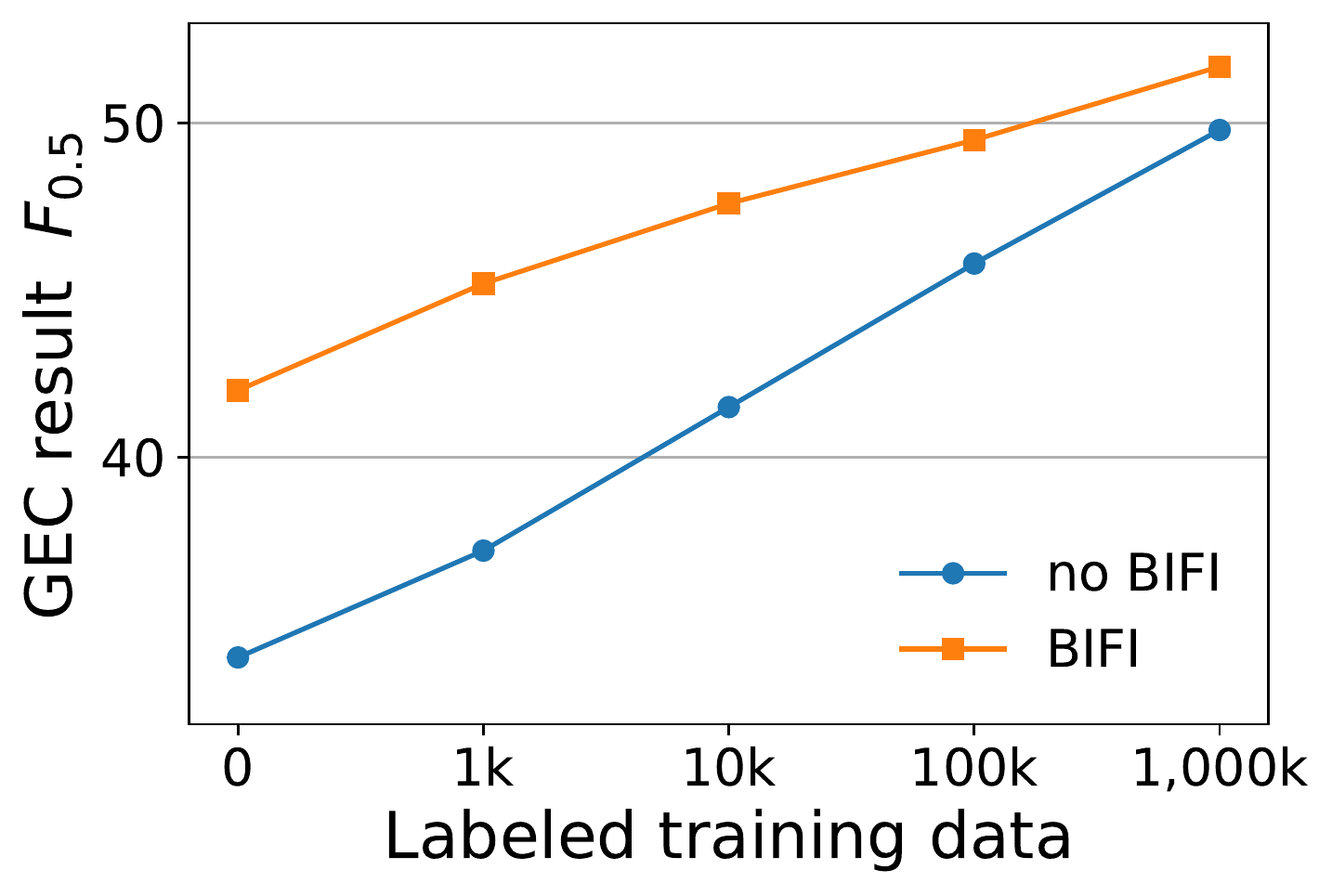}
        \vspace{-3mm}
    \end{center}
    \captionof{figure}{\small
    GEC results ($y$-axis) when varying the amount of labeled data available for training ($x$-axis). BIFI is particularly helpful in low-resource regimes.
    } 
  \label{fig:learning_curve}
\end{minipage}
\end{table*}

\subsection{Experiments}
\label{sec:gec_exp}
We study our proposed approach (BIFI with LM-Critic) on GEC benchmarks, in both unsupervised and supervised settings.

\subsubsection{Evaluation data}
We evaluate on four GEC benchmarks, CoNLL-2014 test \cite{ng2014conll}, BEA-2019 dev \slash test \cite{bryant2019bea}, GMEG-yahoo and GMEG-wiki tests \cite{napoles2019enabling}, which span domains of essay \slash learner English, Wikipedia, and Yahoo!\! Answers.
For CoNLL-2014, we use the official M$^2$ scorer \cite{dahlmeier2012better}, and for others we use the ERRANT metric \cite{bryant2017automatic}. 
We describe the training data separately for unsupervised (\S \ref{sec:gec_exp_low_resource}) and supervised (\S \ref{sec:gec_exp_full_resource}) settings.

\subsubsection{Unsupervised setting}
\label{sec:gec_exp_low_resource}
\heading{Setup and data.}
We consider the setup with no labeled training data.
Existing GEC works (\eg, \citet{awasthi2019parallel,omelianchuk2020gector}) prepare synthetic paired data by heuristically corrupting sentences from the One-billion-word corpus \cite{chelba2013one}.
We follow the same procedure, and train an encoder-decoder Transformer \cite{vaswani2017attention} on this synthetic data to be our \textbf{baseline fixer}. The size of the synthetic data is 9M pairs.

We then apply the BIFI training on top of the baseline fixer.
As our \textbf{unlabeled data} to be used for BIFI, we want text that is likely to contain both ungrammatical and grammatical sentences. Hence, we take 10M sentences in total from the Yahoo!\! Answers corpus \cite{zhang2015character} and the Wikipedia histories data \cite{grundkiewicz2014wiked} for which we take sentences prior to revisions.\footnote{This is not paired data, as we only take sentences pre revision, not post revision.}
This unlabeled data is in the domains of two of our benchmarks (GMEG-wiki and GMEG-yahoo) but not of CoNLL-2014 and BEA-2019.

\heading{Implementation details.}
The encoder-decoder Transformer architecture has 12 layers, 16 attention heads and hidden state size of 768. The model parameters are initialized with the BART-base release \cite{lewis2019bart}, and then optimized by Adam \cite{kingma2015adam}, with batch size of 512 sequences, learning rate 0.0001, and gradient clipping 1.0 \cite{pascanu2013difficulty}, on a single GTX Titan X GPU.
For generation, we use beam search with beam size 10.
We run the BIFI algorithm for $K = 1$ round.
The total training time takes 2 days.

\heading{Results.}
Table \ref{tbl:result_unlabeled} shows the results on the four GEC benchmarks.
``Transformers'' is our baseline fixer, trained on the synthetic paired data. Our proposed approach (``+BIFI'') outperforms the baseline by substantial margins across the benchmarks, \eg, +8 \Fscore on GMEG-wiki and yahoo.

Since our method (``+BIFI'') uses more (unlabeled) data than the baseline (``Transformer''), to be fully fair, we also conduct an experiment that controls the amount of training data seen by the model: Specifically, we apply BIFI to the baseline fixer without the critic, \ie, the model sees the same amount of newly-generated paired data as ``+BIFI'' but they are not verified by LM-Critic. This system (``+BIFI with no critic'') did not improve on the baseline much.
These results indicate that the paired data generated by BIFI with LM-Critic is indeed more realistic and helpful than the initial synthetic data or pairs generated without LM-Critic.

The improved results in this unsupervised setting suggest that our approach is especially useful in domains with no labeled GEC data for training (\eg, GMEG-wiki and yahoo; CoNLL-2014 and BEA-2019 have labeled data, which we use in \S \ref{sec:gec_exp_full_resource}).

Our results also suggest that while existing BIFI assumed access to an oracle critic (\ie, compiler), an approximate critic (\ie, LM-Critic) can also help to improve model learning.
Our conjecture is that as long as the LM-Critic is better than random guessing (\eg, {70 \Fscore} as shown in \S \ref{sec:lm_critic_analysis_performance}), it is useful for improving the quality of GEC training data generated in BIFI (Eq \ref{eq:BIFI-Pf}, \ref{eq:BIFI-Pb}), which in turns improves GEC performance. 
An interesting future direction is to use the breaker learned in BIFI (Eq \ref{eq:BIFI-b} for the perturbation function in LM-Critic (\S \ref{sec:lm_critic_implementation}) to further improve the critic, which may in turn help BIFI as well as GEC performance, creating a positive loop of learning.

\subsubsection{Supervised setting}
\label{sec:gec_exp_full_resource}

\heading{Setup and data.}
We also consider the common leaderboard setup that uses labeled training data and evaluates on CoNLL-2014 and BEA-2019.
We take the state-of-the-art model, GECToR \cite{omelianchuk2020gector}, as our \textbf{baseline fixer}. Following \citet{omelianchuk2020gector}, GECToR is first trained on the synthetic paired data described in \S \ref{sec:gec_exp_low_resource}, and is then trained on the labeled data available for the BEA-2019 task, which is the combination of:
\begin{itemize}
    \item[$\bullet$] NUS Corpus of Learner English (NUCLE) 
     \cite{dahlmeier2013building}
    
    \item[$\bullet$] Lang-8 Corpus of Learner English (Lang-8) 
     \cite{mizumoto2011mining,tajiri2012tense}
    
    \item[$\bullet$] FCE dataset 
     \cite{yannakoudakis2011new}
    
    \item[$\bullet$] Write \& Improve + LOCNESS Corpus (W\&I + LOCNESS) \cite{bryant2019bea}
\end{itemize}
They are all in the domain of CoNLL-2014 and BEA-2019 (learner \slash essay English).
The total size of the labeled data is 1M pairs. 

We then apply the BIFI training on top of GECToR. As our \textbf{unlabeled data} to be used for BIFI, we use 10M sentences taken from Yahoo!\! Answers and Wikipedia histories (same as \S \ref{sec:gec_exp_low_resource}).

\heading{Implementation details.}
We use the same hyperparameters and training procedures for GECToR as in \citet{omelianchuk2020gector}.
We run the BIFI algorithm for $K = 1$ round. 
The total training time takes 4 days, on a single GTX Titan X GPU.

\heading{Results.}
Table \ref{tbl:result_leaderboard} shows our results on CoNLL-2014 test and BEA-2019 test, along with existing systems on the leaderboard.
Our approach (``+BIFI'') provides an additional boost over our base model (``GECToR''). This suggests that BIFI with LM-Critic is helpful not only in the unsupervised setting but also when a substantial amount of labeled data (1M pairs) is available.

\subsubsection{Analysis}
\heading{Varying the amount of labeled data.} 
We have studied GEC results when we have no labeled data (\S \ref{sec:gec_exp_low_resource}) and when we use all the labeled data (1M pairs) (\S \ref{sec:gec_exp_full_resource}).
Here we analyze the interpolation.
In Figure \ref{fig:learning_curve}, we show the GEC performance (\Fscore) on the BEA-2019 dev set, when varying the amount of labeled data available for training from 0 to 1M. The blue line indicates a Transformer model first trained on the synthetic data and then trained on the available labeled data, which is our baseline. The orange line indicates that this baseline model is further trained with BIFI. 
We observe that BIFI outperforms the baseline consistently and is particularly helpful in low-resource regimes.

\heading{Pairs generated by BIFI.}
We quantitatively saw in \S \ref{sec:gec_exp_low_resource} that the paired data generated by BIFI is helpful for learning GEC.
Here we provide qualitative examples to compare the paired data generated by (a) synthetic corruption, (b) BIFI without critic, and (c) BIFI with LM-Critic (Table \ref{tbl:pairs_example}). 
We observe that (a) tends to deviate from the type of grammatical errors humans make (\eg, inserting \slash replacing words arbitrarily); (b) tends to have pairs where $\goodsent$ is broken (\eg, the first pair in Table \ref{tbl:pairs_example}(b)) or $\badsent$ is actually grammatical, as pairs are not verified by a critic; and (c) is the most realistic.

\begin{table}[tb]
\definecolor{lightyellow}{HTML}{FFF7BF}
\definecolor{lightred}{HTML}{f4d7d7} 
\definecolor{lightblue}{HTML}{DAE8FC}
\newcommand{\hlyellow}[1]{{\sethlcolor{yellow}\hl{#1}}}
\newcommand{\hlblue}[1]{{\sethlcolor{lightblue}\hl{#1}}}
\centering
\scalebox{0.7}{
\begin{tabular}{l}
\toprule
\textbf{(a)~ Pairs generated by synthetic corruption} \\
\midrule
\hspace{1mm}$\badsent$:\hspace{1.5mm} {\small We look forward \hlyellow{the} to better treatments in the future.}\\[-1mm]
\hspace{1mm}$\goodsent$: {\small We look forward to better treatments in the future.}\\
\hspace{1mm}$\badsent$:\hspace{1.5mm} {\small The president-elect stayed away so as not to \hlyellow{foregin} matters \hlyellow{until} Bush.}\\[-1mm]
\hspace{1mm}$\goodsent$: {\small The president-elect stayed away so as not to complicate matters for Bush.}\\[2mm]
\toprule
\textbf{(b)~ Pairs generated by BIFI \textit{without} LM-Critic} \\
\midrule
\hspace{1mm}$\badsent$:\hspace{1.5mm} {\small If anyone is interested, here's the kink.}\\[-1mm]
\hspace{1mm}$\goodsent$: {\small If anyone is interested, here's the kink\hlyellow{s}.}\\
\hspace{1mm}$\badsent$:\hspace{1.5mm} {\small \small If you can't find a match yourself, horse trader will help\hlyellow{s}.}\\[-1mm]
\hspace{1mm}$\goodsent$: {\small If you can't find a match yourself, horse trader\hlyellow{s} will help.}\\[2mm]
\toprule
\textbf{(c)~ Pairs generated by BIFI with LM-Critic (Ours)} \\
\midrule
\hspace{1mm}$\badsent$:\hspace{1.5mm} {\small First Light is a award-winning novel by Sunil Gangopadhyay.}\\[-1mm]
\hspace{1mm}$\goodsent$: {\small First Light is a\hlyellow{n} award-winning novel by Sunil Gangopadhyay.}\\
\hspace{1mm}$\badsent$:\hspace{1.5mm} {\small Except latter, the rivers are in underground tubes and not visible.}\\[-1mm]
\hspace{1mm}$\goodsent$: {\small Except \hlyellow{for the} latter, the rivers are in underground tubes and 
not visible.}
\\[0.5mm]

\bottomrule
\end{tabular}
}
\vspace{-2mm}
\caption{\small
Examples of paired data generated by (a) synthetic corruption, (b) BIFI without critic, and (c) BIFI with LM-Critic. 
(a) tends to deviate from the type of grammatical errors humans make. (b) tends to have pairs where $\goodsent$ is broken (\eg, the first pair) or $\badsent$ is already grammatical, as pairs are not verified by a critic. (c) is the most realistic. 
}
\label{tbl:pairs_example}
\end{table}

\begin{table}[tb]
\definecolor{lightyellow}{HTML}{FFF7BF}
\definecolor{lightred}{HTML}{f4d7d7} 
\definecolor{lightblue}{HTML}{DAE8FC}
\newcommand{\hlyellow}[1]{{\sethlcolor{yellow}\hl{#1}}}
\newcommand{\hlblue}[1]{{\sethlcolor{lightblue}\hl{#1}}}
\centering
\scalebox{0.7}{
\begin{tabular}{l}
\toprule
\hspace{-1mm}{\small\bf (Input)}\hspace{0mm} {\small The system is designed to use amplitude comparision for height finding.}\!\!\\[-1mm]
\hspace{-1mm}{\small\bf (Baseline)} {\small The system is designed to use amplitude compa\hlyellow{rison} for height \hlyellow{find}.}\\[-1mm]
\hspace{-1mm}{\small\bf (BIFI)}\hspace{0mm} {\small The system is designed to use amplitude compa\hlyellow{rison} for height finding.}\\
\midrule
\hspace{-1mm}{\small\bf (Input)}\hspace{0mm} {\small Lugu Lake, set in the subalpine zone in Hengduan is a landscape of}\!\!\\[-1.6mm]
\hspace{9.4mm}{\small pine-covered ecoregion.}\\[-1mm]
\hspace{-1mm}{\small\bf (Baseline)} {\small Lugu Lake, set in the subalpine zone in Hengduan\hlyellow{,} is \hlyellow{their} landscape of}\\[-1.6mm]
\hspace{13.2mm}{\small pine-covered ecoregion.}\\[-1mm]
\hspace{-1mm}{\small\bf (BIFI)}\hspace{0mm} {\small Lugu Lake, set in the subalpine zone in Hengduan\hlyellow{,} is a landscape of}\\[-1.6mm]
\hspace{8.4mm}{\small pine-covered ecoregion.}\\
\bottomrule
\end{tabular}
}
\vspace{-2mm}
\caption{\small
Examples where the baseline fixer trained with synthetic data fails but BIFI succeeds. The baseline tends to make unnecessary edits (\eg, changing verb inflection or articles, due to heuristics used when generating synthetic data).
}
\label{tbl:output_example}
\end{table}

\heading{GEC model outputs.}
In Table \ref{tbl:output_example}, we analyze examples where the baseline fixer trained on synthetic data (``Transformer'') fails but our model (``+BIFI'') succeeds. 
We find that the baseline tends to make unnecessary edits (e.g., changing verb inflection or articles), due to the heuristics used when generating synthetic data. In contrast, BIFI achieves higher precision.
\section{Related work and discussion}

\heading{Grammatical error correction (GEC).}
GEC models are commonly trained from human-labeled data \cite{nicholls2003cambridge,dahlmeier2013building,yannakoudakis2011new,bryant2019bea}, or
synthetic data generated by heuristically corrupting unlabeled sentences \cite{awasthi2019parallel,zhao2019improving,grundkiewicz2019neural,katsumata2019almost,omelianchuk2020gector}.
Several works aim to improve the methods for generating paired data, such as learning a breaker from existing labeled data \cite{lichtarge2019corpora}, 
applying backtranslation \cite{sennrich2015improving} to GEC \cite{xie2018noising,kiyono2019empirical}, and synthesizing extra paired data by comparing model predictions and references \cite{ge2018fluency}.
Different from the above works, our method (i) does not require labeled data (works for both unsupervised and supervised settings), and (ii) uses LM-Critic to filter the ``bad''-side and ``good''-side of generated pairs.

\paragraph{Automatic text evaluation.}
Popular metrics used to assess the quality of text in GEC include
GLEU \cite{napoles2015ground,napoles2017jfleg}, M$^2$ \cite{dahlmeier2012better}, ERRANT \cite{bryant2017automatic} and I-measure \cite{felice2015towards}. While these methods require reference text to compare to, LM-Critic does not.
Several prior works also study reference-less methods to assess grammaticality of text:
\citet{wan2005searching,mutton2007gleu,vadlapudi2010automated} use part-of-speech (POS) tagger or parser predictions to score grammaticality;
\citet{napoles2016there,warstadt2018neural,katinskaia2019multiple,niu2020grammaticality} train grammatical error detection (GED) or acceptability judgement systems.
However, these works require POS taggers, parsers or GED systems trained on labeled data, which may not scale or generalize well beyond the domain of training data.
In contrast, LM-Critic only requires an LM, which is unsupervised and can be pretrained on various domains of unlabeled corpora.

\paragraph{Pretrained LM for text evaluation.}
Several works use pretrained LMs for text evaluation.
For reference-based metrics, \citet{zhang2019bertscore} use an LM's embeddings to measure the similarity between input text and reference text.
For reference-less metrics, several works \cite{kann2018sentence,stahlberg2019neural} use an LM's probability as a fluency score of text.
While this provides a continuous score for fluency, it in itself cannot classify grammatical \slash ungrammatical sentences.
Our LM-Critic goes a step further to consider the local optimum criterion for classifying grammaticality.
The reason we want a classifier (critic) is that we work on unsupervised learning of GEC. In the unsupervised setting, there is a distributional shift problem---the synthetically-generated paired data does not match the distribution of grammatical errors humans make. BIFI is a solution for obtaining realistic paired data in an unsupervised way, but it requires a critic. This led us to design a critic for GEC in this work. 
We note that LM-Critic is not meant to replace existing evaluation metrics for GEC, but rather is an approximate critic to assess grammaticality and help the learning of GEC.

Separately, several works \cite{tenney2019bert, hewitt2019structural, yasunaga2019topiceq, cao2020unsupervised} induce grammar or syntactic structures from LMs, suggesting that LMs can learn about grammaticality in an unsupervised way. 
As this capacity is likely to grow with the size of LMs \cite{radford2019language,brown2020language,kaplan2020scaling}, we think that how to leverage pretrained LMs for GEC will become an increasingly important research problem.
\section{Conclusion}
We presented LM-Critic, a method that uses a pretrained language model (LM) as a critic for assessing sentence grammaticality.
Using LM-Critic and the BIFI algorithm, we learn grammatical error correction (GEC) by generating realistic training data from unlabeled text. 
Notably, our approach does not require labeled data, and can also be viewed as an unsupervised method to turn a (GPT2-scale) pretrained LM into an actual GEC system.
Using multiple GEC datasets, we showed that our approach achieves strong performance on unsupervised GEC, suggesting the promise of our method for domains and languages with no labeled GEC data.
We hope this work opens up research avenues in LM-based critics and unsupervised GEC.

\section*{Acknowledgments}
We thank Pang Wei Koh, Tianyi Zhang, Rodrigo Castellon, members of the Stanford P-Lambda, SNAP and NLP groups, as well as our anonymous reviewers for valuable feedback.
This work was supported in part by a Funai Foundation Scholarship and NSF CAREER Award IIS-1552635.

\section*{Reproducibility}
\renewcommand\ttdefault{cmtt}
Code and data are available at\\
\url{https://github.com/michiyasunaga/LM-Critic}.\\
Experiments are available at\\
\url{https://worksheets.codalab.org/worksheets/0x94456a63e1ee4ccfaabdc7f6a356cc82}.

\bibliography{main}
\bibliographystyle{acl_natbib}

\end{document}